\documentclass[10pt,twocolumn,letterpaper]{article}

\usepackage{cvpr}
\usepackage{times}
\usepackage{epsfig}
\usepackage{graphicx}
\usepackage{amsmath}
\usepackage{amssymb}

\usepackage[breaklinks=true,bookmarks=false]{hyperref}

\cvprfinalcopy 

\def\cvprPaperID{5669} 

\ifcvprfinal\fi
\begin{document}

\title{Pointing Novel Objects in Image Captioning\thanks{{\small This work was performed at JD AI Research.}}}

\author{Yehao Li $^{\dag}$$^{\S}$, Ting Yao $^{\ddag}$, Yingwei Pan $^{\ddag}$, Hongyang Chao $^{\dag}$$^{\S}$, and Tao Mei $^{\ddag}$ \\
{\small\centering$^{\dag}$ School of Data and Computer Science, Sun Yat-sen University, Guangzhou, China}\\
{\small\centering$^{\ddag}$ JD AI Research, Beijing, China}\\
{\small\centering$^{\S}$ Key Laboratory of Machine Intelligence and Advanced Computing (Sun Yat-sen University), Ministry of Education}\\
{\tt\small \{yehaoli.sysu, tingyao.ustc, panyw.ustc\}@gmail.com, isschhy@mail.sysu.edu.cn, tmei@live.com}
}

\maketitle
\thispagestyle{empty}


\begin{abstract}
Image captioning has received significant attention with remarkable improvements in recent advances. Nevertheless, images in the wild encapsulate rich knowledge and cannot be sufficiently described with models built on image-caption pairs containing only in-domain objects. In this paper, we propose to address the problem by augmenting standard deep captioning architectures with object learners. Specifically, we present Long Short-Term Memory with Pointing (LSTM-P) --- a new architecture that facilitates vocabulary expansion and produces novel objects via pointing mechanism. Technically, object learners are initially pre-trained on available object recognition data. Pointing in LSTM-P then balances the probability between generating a word through LSTM and copying a word from the recognized objects at each time step in decoder stage. Furthermore, our captioning encourages global coverage of objects in the sentence. Extensive experiments are conducted on both held-out COCO image captioning and ImageNet datasets for describing novel objects, and superior results are reported when comparing to state-of-the-art approaches. More remarkably, we obtain an average of 60.9\% in F1 score on held-out COCO~dataset.
\end{abstract}

\section{Introduction}

Automatic caption generation is the task of producing a natural-language utterance (usually a sentence) that describes the visual content of an image. Practical applications of automatic caption generation include leveraging descriptions for image indexing or retrieval, and helping those with visual impairments by transforming visual signals into information that can be communicated via text-to-speech technology. Recently, state-of-the-art image captioning methods tend to be monolithic deep models essentially of ``encoder-decoder" style \cite{Xiong2016MetaMind,Vinyals14,You:CVPR16}. In general, a Convolutional Neural Network (CNN) is employed to encode an image into a feature vector, and a caption is then decoded from this vector using a Long Short-Term Memory (LSTM) Network, which is one typical Recurrent Neural Network (RNN). Such models have indeed demonstrated promising results on image captioning task. However, one of the most critical limitations is that the existing models are often built on a number of image-caption pairs, which contain only a shallow view of in-domain objects. That hinders the generalization of these models in real-world scenarios to describe novel scenes or objects in out-of-domain images.

The difficulty of novel objects prediction in captioning mainly originates from two aspects: 1) how to facilitate word vocabulary expansion? 2) how to learn a hybrid network that can nicely integrate the recognized objects (words) into the output captions? We propose to mitigate the first problem through leveraging the knowledge from visual recognition datasets, which are freely available and easier to be scalable for developing object learners. Next, \emph{pointing} mechanism is devised to balance the word generation from decoder and the word taken directly from the learnt objects. In other words, such mechanism controls when to directly put the learnt objects at proper places in the output sentence, i.e., \emph{when to point}. Moreover, despite having high quantitative scores, qualitative analysis shows that automatically generated captions by deep captioning models are often limited to describing very generic information of objects, or rely on prior information and correlations from training examples, and resulting frequently in undesired effects such as object hallucination \cite{lu2018neural}. As a result, we further take the coverage of objects into account to cover more objects in the sentence and thus improve the~captions.

By consolidating the idea of pointing mechanism and the coverage of objects into image captioning, we present a new Long Short-Term Memory with Pointing (LSTM-P) architecture for novel object captioning. Given an image, a CNN is utilized to extract visual features, which are fed into LSTM at the initial time step as a trigger of sentence generation. The output of LSTM is probability distribution over all the words in the vocabulary. The pre-trained object recognizers are employed in parallel to detect objects in the input image. A Copying layer then takes the prediction scores of objects and the current hidden state of LSTM as its inputs. It outputs the probability distribution of being copied over all the recognized objects. To dynamically accommodate word generation through LSTM and word copying from the learnt objects, pointing mechanism as a multi-layer perceptron is exploited to balance the output probability distribution from LSTM and copying layer at each time step. Moreover, the coverage of objects is encouraged to talk about more objects found in the image, which is independent of the position in the sentence. As such, the measure of coverage is performed on the bag-of-objects on sentence level. The whole LSTM-P is trained by jointly minimizing the widely-adopted sequential loss on the produced sentence plus sentence-level coverage loss.

The main contribution of this work is the proposal of LSTM-P architecture for addressing the issue of novel objects prediction in image captioning. This issue also leads to the elegant view of how to expand vocabulary, and how to nicely point towards the placements and moments of copying novel objects in the sentence, which are problems not yet fully understood in the literature.

\section{Related Work}\label{sec:RW}

\textbf{Image Captioning.} Inspired from deep learning \cite{Alex:NIPS12} in computer vision and sequence modeling \cite{Sutskever:NIPS14} in Natural Language Processing, modern image captioning methods \cite{Donahue14,rennie2017self,Vinyals14,Xu:ICML15,yao2018exploring,yao2017boosting,You:CVPR16} mainly exploit sequence learning models to produce sentences with flexible syntactical structures. For example, \cite{Vinyals14} presents an end-to-end CNN plus RNN architecture which capitalizes on LSTM to generate sentences word-by-word. \cite{Xu:ICML15} further extends \cite{Vinyals14} by integrating soft/hard attention mechanism to automatically focus on salient regions within images when producing corresponding words. Moreover, instead of calculating visual attention over image regions at each time step of decoding stage, \cite{Xiong2016MetaMind} devises an adaptive attention mechanism in encoder-decoder architecture to additionally decide when to rely on visual signals or language model. Recently, \cite{Wu:CVPR16,yao2017boosting} verify the effectiveness of injecting semantic attributes into CNN plus RNN model for image captioning. Moreover, \cite{You:CVPR16} utilize the semantic attention measured over attributes to boost image captioning. Most recently, \cite{anderson2017bottom} proposes a novel attention based captioning model which exploits object-level attention to enhance sentence generation via bottom-up and top-down attention mechanism.

\textbf{Novel Object Captioning.} The task of novel object captioning has received increasing attention most recently, which leverages additional image-sentence paired data \cite{Mao:ICCV15} or unpaired image/text data \cite{Hendricks:CVPR16,venugopalan2016captioning} to describe novel objects. Existing works mainly remould the RNN-based image captioning frameworks towards the scenario of novel object captioning by additionally leveraging image taggers/object detectors to inject novel objects for~describing. Specifically, \cite{Mao:ICCV15} is one of early attempts that describes novel objects by enlarging the original limited vocabulary based on only a few paired image-sentence data. A transposed weight sharing strategy is especially devised to avoid extensive re-training. In contrast, \cite{Hendricks:CVPR16} presents Deep Compositional Captioner (DCC) which utilizes the largely available unpaired image and text data (e.g., ImageNet and Wikipedia) to facilitate novel object captioning. The knowledge of semantically related objects is explicitly exploited in DCC to compose the sentences containing novel objects. \cite{venugopalan2016captioning} further extends DCC by simultaneously optimizing the visual recognition network, language model, and image captioning network in an end-to-end manner. Recently, \cite{yao2017novel} integrates the regular RNN-based decoder with copying mechanism which can simultaneously copy the detected novel objects to the output sentence. Another two-stage system is proposed in \cite{mogadala2017describing} by firstly building a multi-entity-label image recognition model for predicting abstract concepts and then leveraging such concepts as an external semantic attention \& constrained inference for sentence generation. Furthermore, Anderson \emph{et al.} \cite{anderson2017guided} devise constrained beam search to force the inclusion of selected tag words in the output of RNN-based decoder, facilitating vocabulary expansion to novel objects without re-training. Most recently, \cite{lu2018neural} first generates a hybrid template that contains a mix of words and slots explicitly associated with image region, and then fills in the slots with visual concepts identified in the regions by object~detectors.

\textbf{Summary.} In short, our approach focuses on the latter scenario, that leverages object recognition data for novel object captioning. Similar to previous approaches \cite{mogadala2017describing,yao2017novel}, LSTM-P augments the standard RNN-based language model with the object learners pre-trained on object recognition data. The novelty is on the exploitation of pointing mechanism for dynamically accommodating word generation via RNN-based language model and word copying from the learnt objects. In particular, we utilize the pointing mechanism to elegantly point when to copy the novel objects to target sentence, targeting for balancing the influence between copying mechanism and standard word-by-word sentence generation conditioned on the contexts. Moreover, the measure of sentence-level coverage is adopted as an additional training target to encourage the global coverage of objects in the sentence.

\begin{figure*}
\centering {\includegraphics[width=0.96\textwidth]{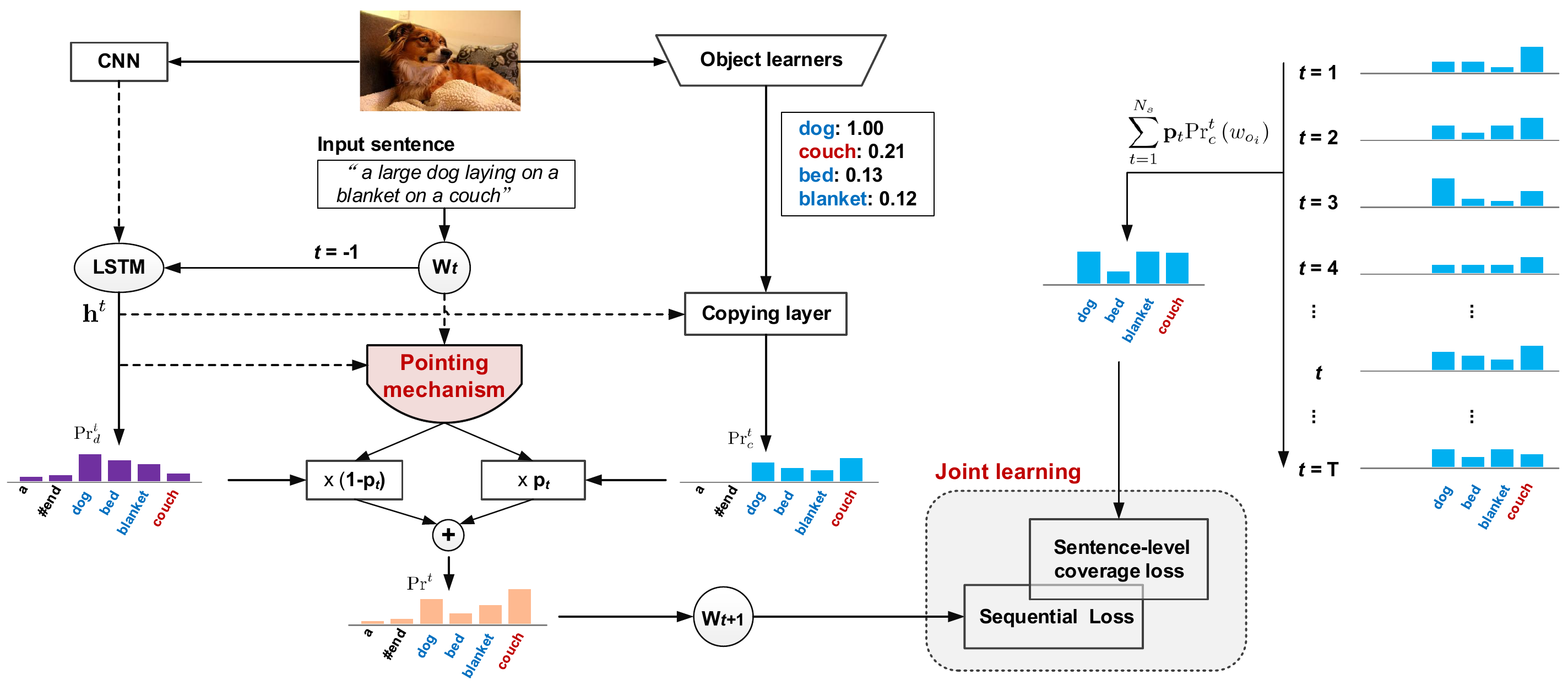}}
\vspace{-0.10in}
\caption{\small An overview of our Long Short-Term Memory with Pointing (LSTM-P) for novel object captioning (better viewed in color). The image representation extracted by CNN is firstly injected into LSTM at the initial time for triggering the standard word-by-word sentence generation. The output of LSTM is the probability distribution over all the words in the vocabulary at each decoding time. Meanwhile, the object learners pre-trained on object recognition data are utilized to detect the objects within the input image. Such predicted score distribution over objects are further injected into a copying layer along with the current hidden state of LSTM, producing the probability distribution of being copied over the recognized objects. To dynamically accommodate word generation via LSTM and word copying from learnt objects, a pointing mechanism is specially devised to elegantly point when to copy the object depending on contextual information (i.e., current input word and LSTM hidden state). The whole LSTM-P is trained by minimizing two objectives in an end-to-end manner: (1) the widely-adopted sequential loss that enforces the syntactic coherence of output sentence, and (2) the sentence-level coverage loss that encourages the maximum coverage of all objects found in the image, which is independent of the position in the sentence.
}
\label{fig:figPC}
\vspace{-0.20in}
\end{figure*}

\section{Method}
We devise our Long Short-Term Memory with Pointing (LSTM-P) architecture to facilitate novel object captioning by dynamically integrating the recognized novel objects into the output sentence via pointing mechanism. In particular, LSTM-P firstly utilizes a regular CNN plus RNN language model to exploit the contextual relationships among the generated words. Meanwhile, the object learners trained on object recognition data are leveraged to detect objects for the input image and a copying layer is further adopted to directly copy a word from the recognized objects. Next, the two pathways for generating target word, i.e., the standard word-by-word sentence generation and the direct copying from recognized objects, are dynamically accommodated through the pointing mechanism, which can point when to copy the novel objects to target sentence conditioned on the context. The overall training of LSTM-P is performed by  simultaneously minimizing the sequential loss that enforces the syntactic coherence of output sentence, and the sentence-level coverage loss that encourages the maximum coverage of all objects found in the image. An overview of our framework is illustrated in Figure \ref{fig:figPC}.

\subsection{Notation}
For novel object captioning task, we aim to describe an input image ${I}$ with a textual sentence $\mathcal{S} = \{w_1, w_2, ..., w_{N_s}\}$ which consists of $N_s$ words. Note that we represent each image ${I}$ as the $D_v$-dimensional visual feature ${\bf{I}}\in {\mathbb{R}}^{D_v}$. Moreover, ${\bf{w}}_t\in {{\mathbb{R}}^{D_w}}$ denotes the $D_w$-dimensional textual feature of the $t$-th word in sentence $\mathcal{S}$. Let $\mathcal {W}_d$ denote the vocabulary on the paired image-sentence data. Furthermore, we leverage the freely available visual recognition datasets to develop the object learners which will be integrated into standard deep captioning architecture for novel object captioning. We denote the object vocabulary for the object recognition dataset as $\mathcal {W}_c$, and ${\bf{I}}_c\in {\mathbb{R}}^{D_c}$ represents the probability distribution over all the $D_c$ objects in $\mathcal {W}_c$ for image $I$ via object learners. Hence the whole vocabulary for our system is denoted as $\mathcal {W}=\mathcal {W}_d \cup \mathcal {W}_c$. In addition, to facilitate the additional measure of object coverage in the sentence, we distill all the objects in textual sentence $\mathcal {S}$ as another training target, denoted as the bag-of-objects $\mathcal{O} = \{w_{o_1}, w_{o_2}, ..., w_{o_{K}}\}$ with $K$ object words.

\subsection{Problem Formulation}
In the novel object captioning problem, on one hand, the words in the sentence should be organized coherently in language, and on the hand, the generated descriptive sentence must be able to address all the objects within image. As such, we can formulate the novel object captioning problem by minimizing the following energy loss function:
\begin{equation}\label{Eq:Eq1}
E(I, \mathcal{S}) = E_d(I, \mathcal{S}) + \lambda \times E_c(I, \mathcal{O}),
\end{equation}
where $\lambda$ is the tradeoff parameter, $E_d(I, \mathcal{S})$ and $E_c(I, \mathcal{O})$ are the sequential loss and sentence-level coverage loss, respectively. The former measures the contextual dependency among the generated sequential words in the sentence through a CNN plus RNN language model which is introduced below. The latter estimates the coverage degree of all objects within image for output sentence, which is presented in Section \ref{sec:CM}.

Specifically, inspired from the sequence learning models in image/video captioning \cite{Donahue14,li2018jointly,pan2016jointly,pan2017video,Vinyals14,Xu:ICML15,yao2017msr} and copying mechanism \cite{yao2017novel}, we equip the regular CNN plus RNN language model with the copying layer, which predicts each target word through not only the word-by-word generation by LSTM-based decoder, but also the direct copying from the recognized objects via copying layer. Hence, the sequential loss $E_d(I, \mathcal{S})$ can be measured as the negative log probability of the correct textual sentence given the image and recognized objects:
\begin{equation}\label{Eq:Eq2}
E_d(I, \mathcal{S})=-\log {\Pr{({\mathcal {S}}|{\bf{I}}, {{\bf{I}}_c})}}.
\end{equation}
As the whole captioning model generates sentence word-by-word, we directly apply chain rule to model the joint probability over the sequential words. Therefore, the $\log$ probability of the sentence is calculated as the sum of the $\log$ probabilities over target words:
\begin{equation}\label{Eq:Eq3}\small
\log {\Pr{({\mathcal {S}}|{\bf{I}}, {{\bf{I}}_c})}} =  \sum\limits_{t = 1}^{{N_s}} {\log {\Pr}^t\left( {\left. {{{\bf{w}}_t}} \right|{\bf{I}}, {{\bf{I}}_c},{{\bf{w}}_0}, \ldots ,{{\bf{w}}_{t - 1}}} \right)}.
\end{equation}
Here the probability of each target word ${\Pr}^t\left( {{\bf{w}}_t}  \right)$ is measured depending on both the probability distribution over the whole vocabulary from LSTM-based decoder and the probability distribution of being copied over the recognized objects from copying layer. To dynamically integrate the influence of such two different probability distributions, we devise a pointing mechanism to adaptively make the decision of which score distribution to focus at each time step, which will be elaborated in Section \ref{sec:PM}.

\subsection{Pointing Mechanism}\label{sec:PM}
When humans have a limited information on how to call an object of interest, it seems natural for humans (and also some primates) to have an efficient behavioral mechanism by drawing attention to objects of interest, i.e., \emph{Pointing} \cite{matthews2012origins}. Such pointing behavior plays the major role in the information delivery and can naturally associate context to a particular object without knowing how to call it, i.e., the novel object that never seen before. Inspired from the pointing behavior and the pointer networks \cite{vinyals2015pointer}, we design a pointing mechanism to deal with the novel objects in image captioning scenario. More precisely, the pointing mechanism is a hybrid between the regular LSTM-based language model plus a copying layer and a pointing behavior. It facilitates directly copying recognized objects, which concentrates on the handling of novel objects, while retraining the ability to generate coherence words in grammar via language model. The interactions between LSTM plus copying layer and the pointing mechanism is depicted in the left part of Figure \ref{fig:figPC}.

Specifically, in the decoding stage, given the current LSTM cell output ${\bf{h}}^{t}$ at the $t$-th time step, two probability distributions over the whole vocabulary $\mathcal {W}$ and the object vocabulary $\mathcal {W}_c$ are firstly calculated with regard to the regular sequence modeling in LSTM and the direct copying of objects in copying layer, respectively. For the probability distribution over the whole vocabulary of LSTM, the corresponding probability of generating any target word $w_{t+1} \in \mathcal {W}$ is measured as
\begin{equation}\label{Eq:Eq4}
{\Pr}^t_d\left( {{w_{t + 1}}} \right) = {{\bf{w}}_{t+1}^\top}{{\bf{M}}_d}{{\bf{h}}^t},
\end{equation}
where $D_h$ is the dimensionality of LSTM output and ${{\bf{M}}_d} \in {{\mathbb{R}}^{{D_w} \times {D_h}}}$ is the transformation matrix for textual features of word. For the probability distribution of being copied over the object vocabulary, we directly achieve the probability of copying any object $w_{t+1} \in \mathcal {W}_c$ conditioned on the current LSTM cell output ${\bf{h}}^{t}$ and the output of object learners ${{\bf{I}}_c}$:
\begin{equation}\label{Eq:Eq5}
{\Pr}^t_c\left( {{w_{t + 1}}} \right) ={{\bf{w}}_{t + 1}^\top{{\bf{M}}^1_c}}  \left( {{{\bf{I}}_c}\odot\sigma\left({{\bf{M}}^2_c}{\bf{h}}^t\right)}\right),
\end{equation}
where ${{\bf{M}}^1_c} \in {{\mathbb{R}}^{{D_w} \times {D_c}}}$ and ${{\bf{M}}^2_c} \in {{\mathbb{R}}^{{D_c} \times {D_h}}}$ are the transformation matrices, $\sigma$ is the sigmoid function and $\odot$ is the element-wise dot product function.

Next, the pointing mechanism encapsulates dynamic contextual information (current input word and LSTM cell output) to learn when to point novel objects for copying, which is applied with feature transformation, to produce a weight value and followed by a sigmoid function to squash the weight value to a range of $[0,1]$. Such output weight value ${\bf{p}}_t$ in pointing mechanism is computed as
\begin{equation}\label{Eq:Eq6}
{{\bf{p}}_t} = \sigma ({{\bf{G}}_s}{{\bf{w}}_t} + {{\bf{G}}_h}{{\bf{h}}^{t}} + {\bf{b}}_p),
\end{equation}
where ${{\bf{G}}_s} \in {{\mathbb{R}}^{{{D_{w}}}}}$, ${{\bf{G}}_h} \in {{\mathbb{R}}^{ {D_{h}}}}$ are the transformation matrices for textual features of word and cell output of LSTM respectively, and ${\bf{b}}_p$ is the bias. Here the weight value ${\bf{p}}_t$ is adopted as a soft switch to choose between generating a word through LSTM, or directly copying a word from the recognized objects. As such, the final probability of each target word $w_{t+1}$ over the whole vocabulary $\mathcal {W}$ is obtained by dynamically fusing the two probability distributions in Eq.(\ref{Eq:Eq4}) and Eq.(\ref{Eq:Eq5}) with the weight value ${\bf{p}}_t$:
\begin{equation}\label{Eq:Eq7}\small
\begin{split}
{\Pr}^t\left( {w_{t+1}}  \right) = {\bf{p}}^t_d\cdot\phi({\Pr}^t_d\left( {{w_{t + 1}}} \right))+{\bf{p}}^t_c \cdot \phi ( {\Pr}^t_c\left( {{w_{t + 1}}} \right)),\\
{\bf{p}}^t_d=1-{\bf{p}}_t, ~~{\bf{p}}^t_c={\bf{p}}_t,~~~~~~~~~~~~~~~~~~~~~~~~~~~
\end{split}
\end{equation}
where ${\bf{p}}^t_d$ and ${\bf{p}}^t_c$ denote the weight for generating a word via LSTM or copying a word from recognized objects. $\phi$ represents a softmax function.

\subsection{Coverage of Objects}\label{sec:CM}
While high quantitative scores have been achieved through RNN-based image captioning systems in encoder-decoder paradigm, there is increasing evidence \cite{das2016human,lu2018neural} revealing that such paradigm still lacks visual grounding (i.e., do not associate mentioned concepts to pixels of image). As such, the generated captions are more prone to describe generic information of objects or even copy the most frequently phrases/captions in training data, resulting in undesired effects such as object hallucination. Accordingly, we further measure the coverage of objects as additional training target to holistically cover more objects in the sentence, aiming to emphasis the correctness of mentioned objects regardless of syntax structure and thus improve the captions.

In particular, measuring the coverage of objects is formulated as the multi-label classification problem. We firstly accumulate all the probability distributions of being copied on the object vocabulary generated at decoding stage. The normalized sentence-level probability distribution for copying is thus obtained via aggregating all the probability distributions for copying weighted by the weight value ${\bf{p}}_t$ in pointing mechanism, followed by a sigmoid normalization:
\begin{equation}\label{Eq:Eq8}
{\Pr}_s\left( {{w_{o_i}}} \right) = \sigma\left({\sum\limits_{t = 1}^{{N_s}} { {\bf{p}}_t {\Pr}^t_c\left( {{w_{o_i}}} \right)}}\right).
\end{equation}
Here the sentence-level probability for each object ${w_{o_i}} \in \mathcal {W}_c$ represents how possible the object been directly copied in the generated sentence regardless of the position in the sentence. Thus, the sentence-level coverage loss is calculated as the cross entropy loss in multi-label classification:
\begin{equation}\label{Eq:Eq9}
E_c(I, \mathcal{O}) = -\sum\limits_{i = 1}^{{K}} {\log {\Pr}_s\left( {{w_{o_i}}} \right)}.
\end{equation}
By minimizing this sentence-level coverage loss, the captioning system is encouraged to talk about more objects found in the image.

\subsection{Optimization}
\textbf{Training.} The overall training objective of our LSTM-P integrates the widely-adopted sequential loss in Eq.(\ref{Eq:Eq2}) and sentence-level coverage loss in Eq.(\ref{Eq:Eq9}). Hence we obtain the following optimization problem:
\begin{equation}\label{Eq:Eq10}
\mathcal{L} = -\sum\limits_{t = 1}^{{N_s}} {\log {\Pr}^t\left( {{{{\bf{w}}_t}} } \right)} - \lambda \sum\limits_{i = 1}^{{K}} {\log {\Pr}_s\left( {{w_{o_i}}} \right)},
\end{equation}
where $\lambda$ is tradeoff parameter. With this overall loss objective, the crucial goal of this optimization is to encourage the generated sentence to be coherent in language and meanwhile address all the objects within image.

\textbf{Inference.} In the inference stage, we choose output word among the whole vocabulary ${\mathcal {W}}$ with maximum probability at each time step with the guidance from pointing mechanism. The embedded textual feature of output word is set as LSTM input for the next time step. This process continues until the end sign word is emitted or the pre-defined maximum sentence length is reached.

\section{Experiments}\label{sec:EX}
We conduct extensive evaluations of our proposed LSTM-P for novel object captioning task on two image datasets, including the held-out COCO image captioning dataset (\textbf{held-out COCO}) \cite{Hendricks:CVPR16}, a subset of image captioning benchmark---COCO \cite{Lin:ECCV14}, and \textbf{ImageNet} \cite{ILSVRC15} which is a large-scale object recognition dataset.

\subsection{Dataset and Experimental Settings}
\textbf{Dataset.}
The \textbf{held-out COCO} consists of a subset of COCO by excluding all the image-sentence pairs which contain at least one of eight specific objects in COCO: ``bottle," ``bus," ``couch," ``microwave," ``pizza," ``racket," ``suitcase," and ``zebra". In this dataset, each image is annotated with five descriptions by humans. Since the annotations of the official testing set are not publicly available, we follow the split in \cite{Hendricks:CVPR16} and take half of COCO validation set as validation set and the other half for testing. In the experiments, we firstly train the object learners with all the images in COCO training set including the eight novel objects, and the LSTM is pre-trained with all the sentences from COCO training set. Next, all the paired image-sentence data from held-out COCO training set are leveraged to optimize our novel object captioning system. Our LSTM-P model is finally evaluated over the testing set of held-out COCO to verify the ability of describing the eight novel objects.

\textbf{ImageNet} is the large-scale object recognition dataset and we adopt a subset of ImageNet containing 634 objects that are not present in COCO for evaluation, as in \cite{venugopalan2016captioning}. Specifically, we take about 75\% of images in each class for training and the rest for testing. Hence the training and testing sets include 493,519 and 164,820 images in total. In the experiments, we firstly train the object learners with the entire ImageNet training set, and the LSTM is pre-trained with all the sentences from COCO training set. After that, our novel object captioning system is optimized with all the paired image-sentence data from COCO training set. During inference, we directly produce sentences for testing images in ImageNet and evaluate the ability of describing the 634 novel objects for our LSTM-P.

\begin{table*}[!tb]
\centering
\setlength{\tabcolsep}{1.2pt}
\caption{Per-object F1, averaged F1, SPICE, METEOR, and CIDEr scores of our proposed model and other state-of-the-art methods on held-out COCO dataset for novel object captioning. All values are reported as percentage (\%).}
\label{table:FMCOCO}
\begin{tabular}{l|cccccccc|c|ccc}\hline
~~Model&~F1$_\text{bottle}$~&~F1$_\text{bus}$~&~F1$_\text{couch}$~&~F1$_\text{microwave}$~&~F1$_\text{pizza}$~&
~F1$_\text{racket}$~&~F1$_\text{suitcase}$~&~F1$_\text{zebra}$~&~F1$_\text{average}$~&~SPICE~&~METEOR~&~CIDEr~
\\ \hline\hline
~~LRCN \cite{Donahue14}                &  0     &  0     &  0     &  0     &  0     &  0     &  0     &  0     &  0     &  -     & 19.3  &  -     \\\hline
~~DCC \cite{Hendricks:CVPR16}          &  4.6   & 29.8   & 45.9   & 28.1   & 64.6   & 52.2   & 13.2   & 79.9   & 39.8   & 13.4   & 21.0  &  59.1  \\
~~NOC \cite{venugopalan2016captioning} & 14.9   & 69.0   & 43.8   & 37.9   & 66.5   & 65.9   & 28.1   & 88.7   & 51.8   &  -     & 20.7  &  -     \\
~~NBT \cite{lu2018neural}              &  7.1   & 73.7   & 34.4   & 61.9   & 59.9   & 20.2   & 42.3   & 88.5   & 48.5   & 15.7   & 22.8  &  77.0  \\
~~Base+T4 \cite{anderson2017guided}    & 16.3   & 67.8   & 48.2   & 29.7   & 77.2   & 57.1   & 49.9   & 85.7   & 54     & 15.9   & 23.3  &  77.9  \\
~~KGA-CGM \cite{mogadala2017describing}        & 26.4   & 54.2   & 42.1   & 50.9   & 70.8   & 75.3   & 25.6   & 90.7    & 54.5   & 14.6  & 22.2  & -  \\
~~LSTM-C  \cite{yao2017novel}          & 29.7   & 74.4   & 38.8   & 27.8   & 68.2   & 70.3   & 44.8   & 91.4   & 55.7   & -      & 23.0  & -   \\
~~DNOC  \cite{wu2018decoupled}         & 33.0   & 76.9   & 54.0   & 46.6   & 75.8   & 33.0   & 59.5   & 84.6   & 57.9   & -      & 21.6  & -   \\\hline
~~LSTM-P$^{-}$                             & 26.7   & 74.5   & 46.2   & 50.5   & 81.7   & 47.2   & 61.1   & 91.9   & 60.0   & 16.5  &  23.2  & 88.0  \\
~~LSTM-P                       & 28.7   & 75.5   & 47.1  & 51.5  & 81.9  & 47.1 & 62.6 & 93.0 & \textbf{60.9} & \textbf{16.6} & \textbf{23.4} & \textbf{88.3}  \\
\hline
\end{tabular}
\vspace{-0.2in}
\end{table*}

\textbf{Implementation Details.}
For fair comparison with other state-of-the-art methods, we take the output of 4,096-dimensional fc7 from 16-layer VGG \cite{Simonyan14} pre-trained on ImageNet \cite{ILSVRC15} as image representation. Each word in the sentence is represented as Glove embeddings \cite{pennington2014glove}. For the object learners on COCO, we select only the 1,000 most common words from COCO and utilize MIL model \cite{Fang:CVPR15} to train the object learners over the whole training data of COCO. For the object learners on ImageNet, we directly fine-tune 16-layer VGG pre-trained on ImageNet to obtain the 634 object learners. The hidden layer size in LSTM is set as 1,024. The tradeoff parameter $\lambda$ to balance the sequential loss and the sentence-level coverage loss is empirically set to 0.3. Following \cite{venugopalan2016captioning}, we implicitly integrate the overall energy loss with a text-specific loss on external sentence data for maintaining the model's ability to address novel objects among sentences and a binary classification loss to guide the learning of pointing mechanism. Our novel object captioning model is mainly implemented on Caffe \cite{Jia:MM14}, one of widely adopted deep learning frameworks. Specifically, we set the initial learning rate as 0.0005 and the mini-batch size is set as 512. For all experiments, the maximum training iteration is set as 50 epoches.

\textbf{Evaluation Metrics.}
To quantitatively evaluate our LSTM-P on held-out COCO dataset, we utilize the most common metrics of image captioning task, i.e., \textbf{METEOR} \cite{Banerjee:ACL05}, \textbf{CIDEr} \cite{vedantam2015cider}, and \textbf{SPICE} \cite{spice2016}, to evaluate the quality of generated description. In addition, \textbf{F1-score} \cite{Hendricks:CVPR16} is adopted to further evaluate the ability of describing novel objects. Note that the metric of F1-score indicates whether the novel object is addressed in the generated sentences of the given image which contains that novel object. In our experiments, for fair comparison, both of the METEOR and F1-score metrics are calculated by utilizing the codes\footnote{\url {https://github.com/LisaAnne/DCC}} released by \cite{Hendricks:CVPR16}. For the evaluation on ImageNet which contains no ground-truth sentences, we follow \cite{venugopalan2016captioning} and adopt another two metrics: describing novel objects (\textbf{Novel}) and \textbf{Accuracy} scores. Here the Novel score calculates the percentage of all the 634 novel objects addressed in the generated sentences. In other words, for each novel object, the model should mention it within at least one description for the image containing this object. The Accuracy score of each novel object denotes the percentage of images containing this novel object which can be correctly described by mentioning that novel object in generated descriptions. We obtain the final Accuracy score by averaging all the accuracy scores of 634 novel objects.

\subsection{Compared Approaches}
We compare our LSTM-P model with the following state-of-the-art methods, which include both the regular image captioning methods and novel object captioning models: (1) \textbf{LRCN} \cite{Donahue14} is a basic LSTM-based captioning model which triggers sentence generation by injecting input image and previous word into LSTM at each time step. We directly train LRCN on the paired image-sentence data without any novel objects. (2) \textbf{DCC} \cite{Hendricks:CVPR16} leverages external unpaired data to pre-train lexical classifier and language model. Next, the whole captioning framework is trained with paired image-sentence data. (3) \textbf{NOC} \cite{venugopalan2016captioning} presents a novel object captioning system consisting of visual recognition network, LSTM-based language model, and image captioning network. The three components are simultaneously optimized in an end-to-end fashion. (4) \textbf{NBT} \cite{lu2018neural} first generates a hybrid template that contains a mix of words and slots associated with image region, and then fills in the slots with detected visual concepts. (5) \textbf{Base+T4} \cite{anderson2017guided} designs constrained beam search to force the inclusion of predicted tag words in the output of RNN-based decoder without re-training. (6) \textbf{KGA-CGM} \cite{mogadala2017describing} takes the predicted concepts as an external semantic attention and constrained inference for sentence generation. (7) \textbf{LSTM-C} \cite{yao2017novel} integrates the standard RNN-based decoder with copying mechanism which can directly copy the predicted objects into the output sentence. (8) \textbf{DNOC} \cite{wu2018decoupled}  generates the caption template with placeholder and then fill in the placeholder with the detected objects via key-value object memory. (9) \textbf{LSTM-P} is the proposal in this paper. Moreover, a slightly different version of this run is named as \textbf{LSTM-P$^{-}$}, which is trained without the sentence-level coverage loss.

\begin{figure}[!tb]
\centering {\includegraphics[width=0.5\textwidth]{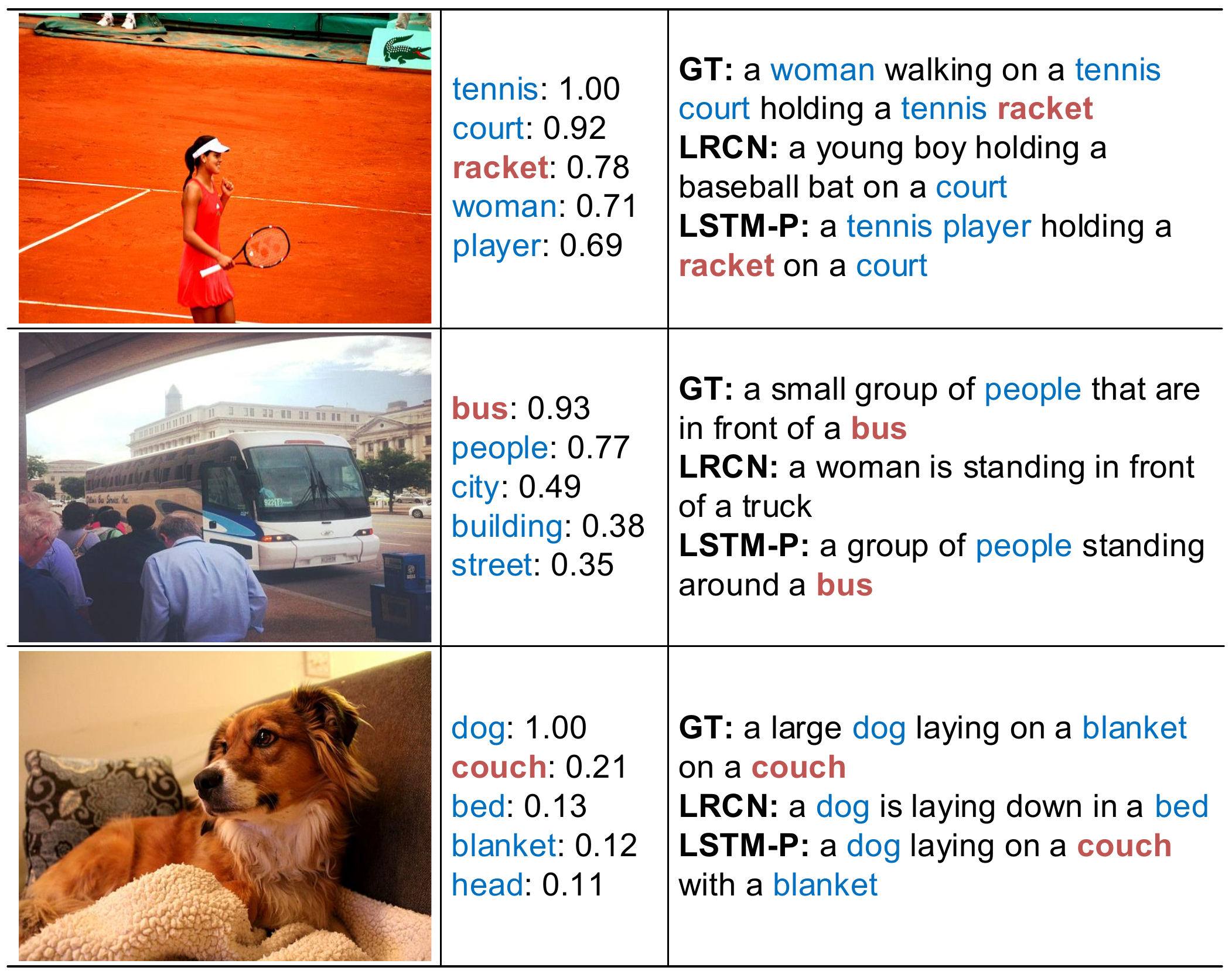}}
\vspace{-0.15in}
\caption{\small Objects and sentence generation results on held-out COCO. The detected objects are predicted by MIL model in \cite{Fang:CVPR15}, and the output sentences are generated by 1) Ground Truth (GT): one ground truth sentence, 2) LRCN and 3) our LSTM-P.}
\label{fig:figRCOCO}
\vspace{-0.15in}
\end{figure}

\subsection{Performance Comparison}
\textbf{Evaluation on held-out COCO.} Table \ref{table:FMCOCO} shows the performances of compared ten models on held-out COCO dataset. Overall, the results across all the four general evaluation metrics consistently indicate that our proposed LSTM-P exhibits better performance than all the state-of-the-art techniques including regular image captioning model (LRCN) and seven novel object captioning systems. In particular, the F1$_\text{average}$ score of our LSTM-P can achieve 60.9\%, making the relative improvement over the best competitor by 5.2\%, which is generally considered as a significant progress on this dataset. As expected, by additionally utilizing external object recognition data for training, all the latter nine novel object captioning models outperform the regular image captioning model LRCN on both description quality and novelty. By augmenting the standard RNN-based language model with the object/concept learners, LSTM-C leads to a performance boost against NOC that produces novel objects purely depending on generative mechanism in LSTM. The results basically indicate that the advantage of directly ``copying" the predicted objects/concepts into output sentence via copying mechanism. However, the performances of LSTM-C are still lower than our LSTM-P$^{-}$, which leverages the pointing mechanism to balance the influence between copying mechanism and standard word-by-word sentence generation conditioned on the contexts. This confirms the effectiveness of elegantly pointing when to copy the novel objects to target sentence for novel object captioning. In addition, by further integrating sentence-level coverage loss into overall training objective, LSTM-P exhibits better performance than LSTM-P$^{-}$, which demonstrates the merit of encouraging the generated sentence to be coherent in language and meanwhile address all the objects within image.

\begin{figure}[!tb]
\centering {\includegraphics[width=0.5\textwidth]{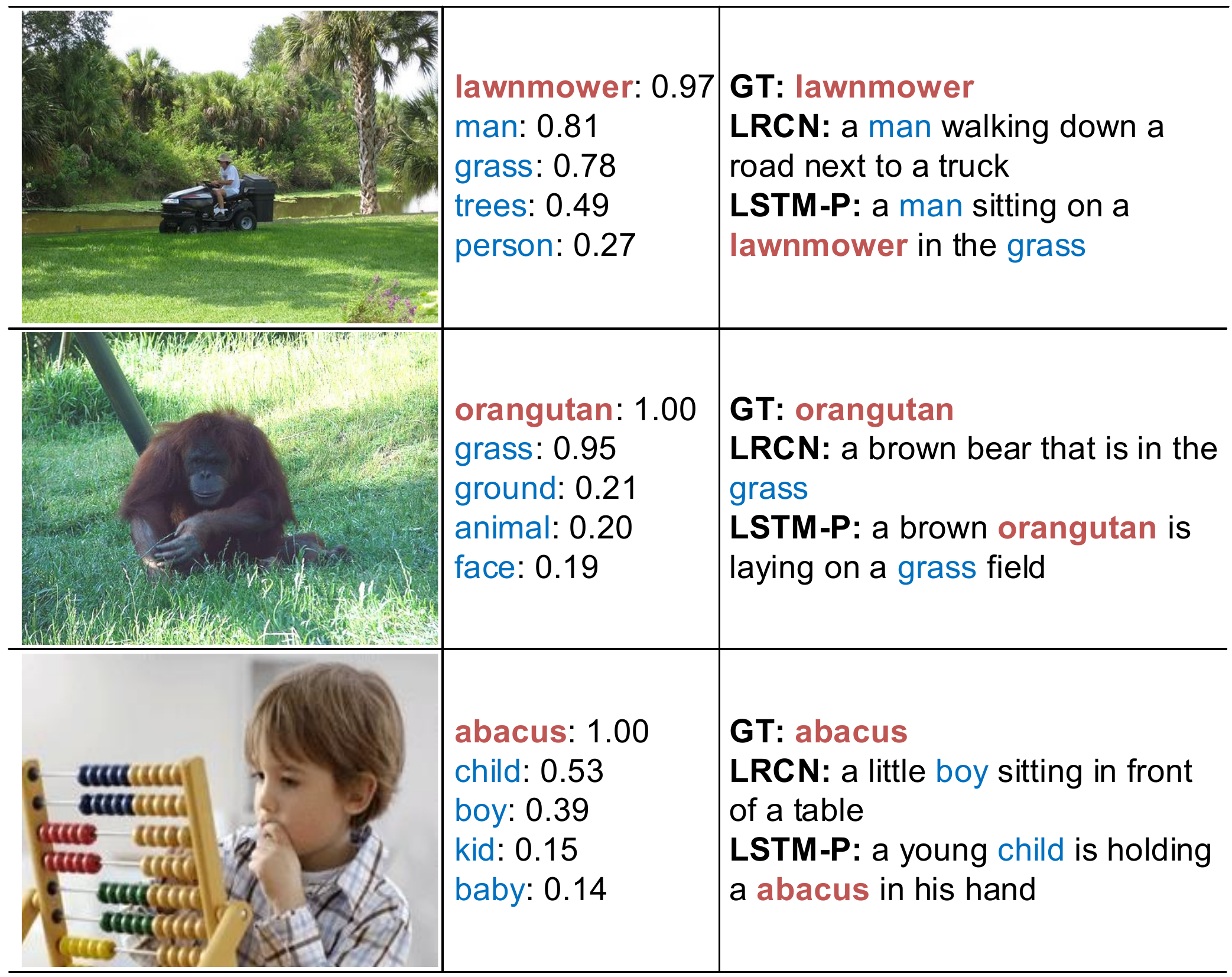}}
\caption{Objects and sentence generation results on ImageNet. GT denotes the ground truth object. The detected objects are predicted by the standard CNN architecture \cite{Simonyan14}, and the output sentences are generated by 1) LRCN and 2) our LSTM-P.}
\label{fig:figImageNet}
\vspace{-0.1in}
\end{figure}

\begin{table}[!tb]\small
\centering
\caption{Novel, F1 and Accuracy scores of our proposed model and other state-of-the-art methods on ImageNet dataset. All values are reported as percentage (\%).}
\label{table:NFIM}
\begin{tabular}{l|c|c|c}\hline
Model~~&~~Novel~~&~~F1~~&~~Accuracy~~
\\ \hline\hline
NOC \cite{venugopalan2016captioning} & & & \\
~~~~-COCO  &69.08 & 15.63 & 10.04\\
~~~~-BNC\&Wiki &87.69  & 31.23 &21.96\\
LSTM-C  \cite{yao2017novel} & & & \\
~~~~-COCO  &72.08 & 16.39 & 11.83\\
~~~~-BNC\&Wiki &89.11  & 33.64 &31.11\\\hline
LSTM-P  & & & \\
~~~~-COCO    & 90.06 & 17.67 & 11.91 \\
~~~~-BNC\&Wiki & 91.17 & 52.07 & 44.63 \\
\hline
\end{tabular}
\vspace{-0.2in}
\end{table}

\textbf{Evaluation on ImageNet.} To further verify the scalability of our proposed LSTM-P, we additionally perform experiment on ImageNet to describe hundreds of novel objects that outside of the paired image-sentence data. Table \ref{table:NFIM} shows the performance comparison on ImageNet. Similar to the observations on held-out COCO, our LSTM-P exhibits better performance than other runs. In particular, the Novel, F1, and Accuracy scores for LSTM-P can reach $90.06\%$, $17.67\%$, and $11.91\%$, making the relative improvement over LSTM-C by $24.9\%$, $7.8\%$, and $0.7\%$, respectively. The results basically indicate the advantage of exploiting pointing mechanism to balance the word generation from decoder and the word copied from learnt objects, and the global coverage of objects in output sentence, for novel object captioning, even when scaling into ImageNet images with hundreds of novel objects. Moreover, we follow \cite{venugopalan2016captioning,yao2017novel} and include the external unpaired text data (i.e., British National Corpus and Wikipedia) for training our LSTM-P. The performance gains are further attained.

\subsection{Experimental Analysis}
In this section, we further analyze the qualitative results, the weights visualization in pointing mechanism, and the effect of the tradeoff parameter $\lambda$ for novel object captioning task on held-out COCO dataset.

\begin{figure*}[!tb]
\vspace{-0.00in}
\centering {\includegraphics[width=0.96\textwidth]{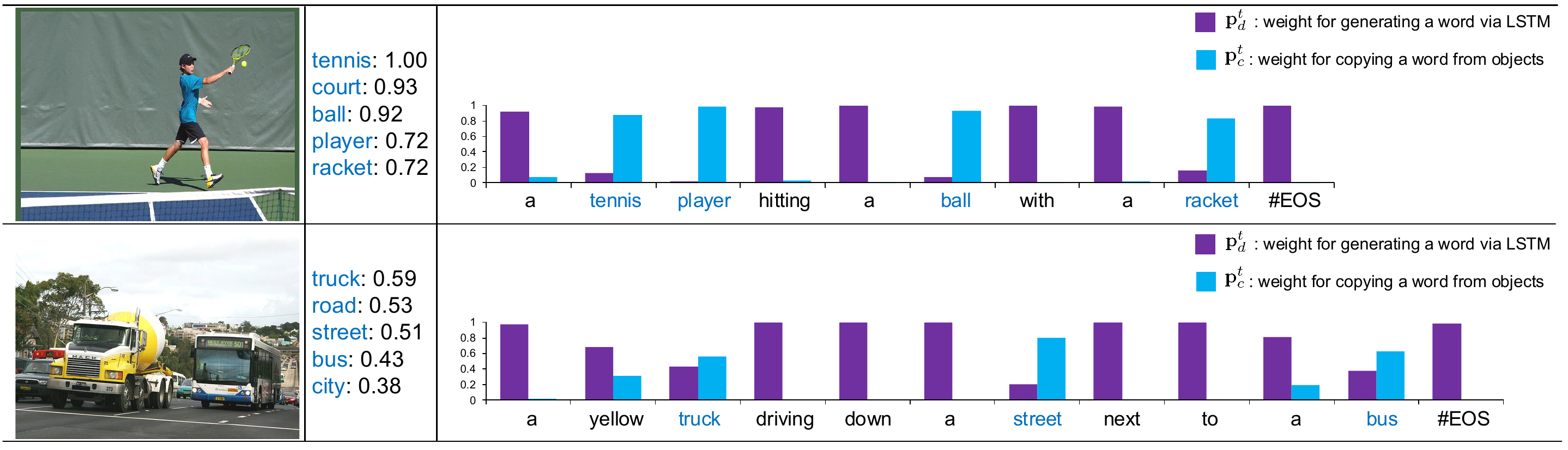}}
\vspace{-0.05in}
\caption{\small Sentence generation results with visualized weights learnt in pointer mechanism of our LSTM-P at each decoding step on held-out COCO dataset. The bar plot at each decoding step corresponds to the weights for generating a word via LSTM or copying a word from recognized objects when the corresponding word was generated.}
\label{fig:figpointer}
\vspace{-0.25in}
\end{figure*}

\textbf{Qualitative Analysis.} Figure \ref{fig:figRCOCO} and Figure \ref{fig:figImageNet} showcase a few sentence examples generated by different methods, the detected objects and human-annotated ground truth on held-out COCO and ImageNet, respectively. From these exemplar results, it is easy to see that all of these captioning models can generate somewhat relevant sentences on both datasets, while our proposed LSTM-P can correctly describe the novel objects by learning to point towards the placements and moments of copying novel objects via pointing mechanism. For example, compared to object term ``bed" in the sentence generated by LRCN, ``couch" in our LSTM-P is more precise to describe the image content in the last image on held-out COCO dataset, since the novel object ``couch" is among the top object candidates and directly copied to the output sentence at the corresponding decoding step. Moreover, by additionally measuring the coverage over the bag-of-objects on sentence level, our LSTM-P is encouraged to produce sentences which cover more objects found in images, leading to more descriptive sentence with object ``blanket."

\begin{figure}[!tb]
\centering {\includegraphics[width=0.5\textwidth]{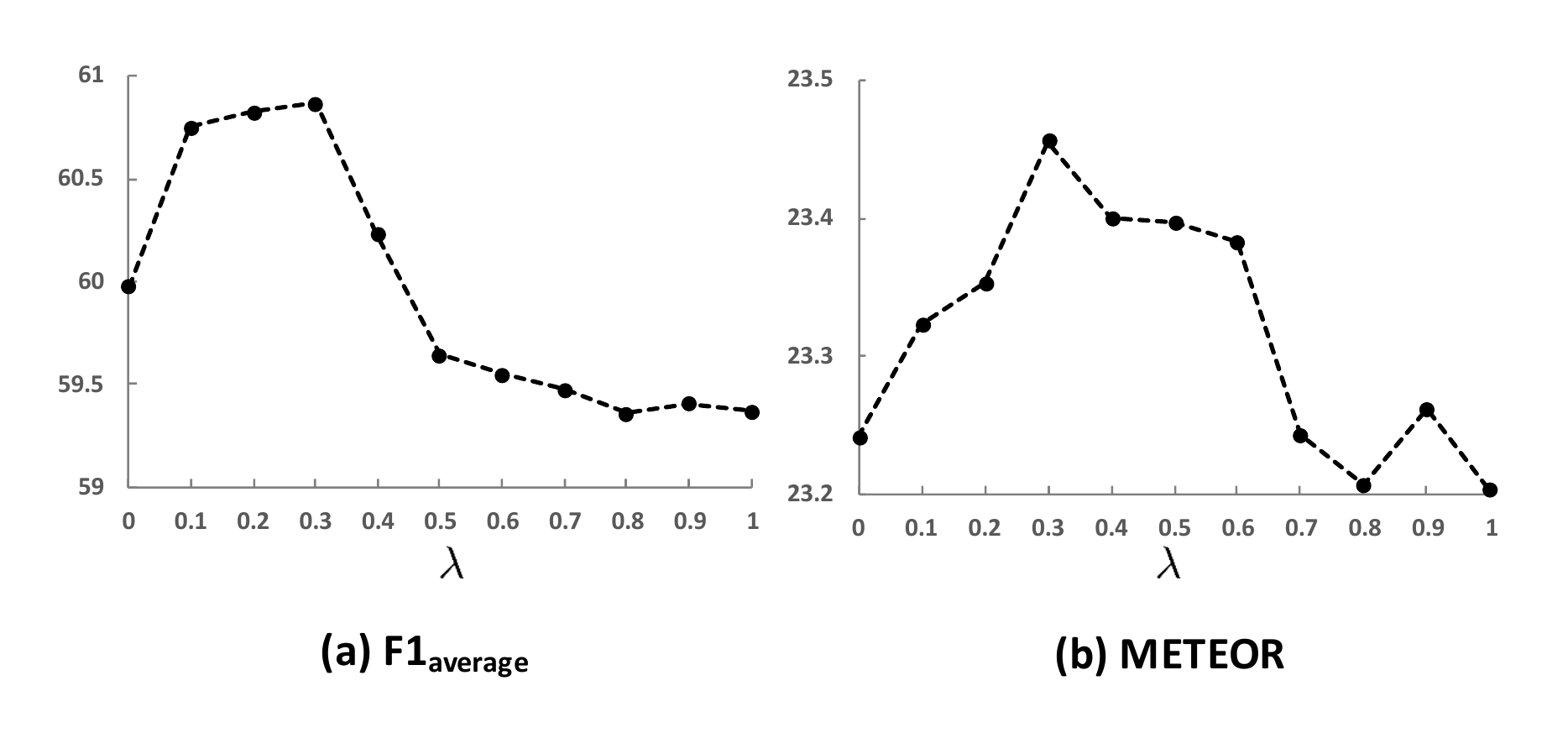}}
\vspace{-0.2in}
\caption{\small The effect of the tradeoff parameter $\lambda$ in our LSTM-P over (a) F1$_\text{average}$ (\%) and (b) METEOR (\%) on held-out COCO.}
\label{fig:figtradeoff}
\vspace{-0.2in}
\end{figure}

\textbf{Visualization of weights in pointing mechanism.} To better qualitatively evaluate the generated results with pointing mechanism of our LSTM-P, we further visualize the generated weights of generating a word via LSTM or copying a word from recognized objects for a few examples in Figure \ref{fig:figpointer}. We can easily observe that our LSTM-P correctly chooses to copy word from recognized objects when the object word to be generated. For instance, in the first image, when LSTM-P is about to generate object word (i.e., ``tennis," ``player," ``ball," and ``racket"), it mostly prefer to copy the object word from recognized objects with higher weight value ${\bf{p}}^t_c$. Also, for the second video, the pointer mechanism attends to direct copying from objects when the object terms (i.e., ``truck," ``street," and ``bus") are about to be generated at decoding stage.

\textbf{Effect of the Tradeoff Parameter $\lambda$.}
To clarify the effect of the tradeoff parameter $\lambda$ in Eq.(\ref{Eq:Eq10}), we illustrate the performance curves over two evaluation metrics with a different tradeoff parameter in Figure \ref{fig:figtradeoff}. As shown in the figure, we can see that all performance curves are generally like the ``$\wedge$" shapes when $\lambda$ varies in a range from 0 to 1. Hence we set the tradeoff parameter $\lambda$ as 0.3 in our experiments, which can achieve the best performance. This again proves that it is reasonable to encourage both the syntactic coherence and the global coverage of objects in the output sentence for boosting novel object captioning.

\section{Conclusions}
We have presented Long Short-Term Memory with Pointing (LSTM-P) architecture which produces novel objects in image captioning via pointing mechanism. Particularly, we study the problems of how to facilitate vocabulary expansion and how to learn a hybrid network that can nicely integrate the recognized objects into the output caption. To verify our claim, we have initially pre-trained object learners on free available object recognition data. Next the pointing mechanism is devised to balance the word generation from RNN-based decoder and the word taken directly from the learnt objects. Moreover, the sentence-level coverage of objects is further exploited to cover more objects in the sentence and thus improve the captions. Experiments conducted on both held-out COCO image captioning and ImageNet datasets validate our model and analysis. More remarkably, we achieve new state-of-the-art performance of single model: 60.9\% in F1 score on held-out COCO~dataset.

\textbf{Acknowledgments.} This work is partially supported by NSF of China under Grant 61672548, U1611461, 61173081, and the Guangzhou Science and Technology Program, China, under Grant 201510010165.

{\small
\bibliographystyle{ieee}
\bibliography{egbib}
}

\end{document}